\documentclass[10pt,twocolumn,letterpaper,pagebackref,breaklinks,colorlinks,allcolors=wacvblue]{article}

\usepackage{wacv}              
\usepackage{graphicx}
\usepackage{booktabs}

\usepackage{orcidlink}
\usepackage{mathtools}
\usepackage{comment}
\usepackage{algorithm}
\usepackage{lipsum}
\usepackage{algpseudocode}

\usepackage{mathtools}
\DeclareMathAlphabet{\mathbbold}{U}{bbold}{m}{n}
\usepackage{amsmath}  
\usepackage{amssymb}  
\usepackage{dsfont}
\usepackage{verbatim}
\usepackage{bbm}
\usepackage{multirow}
\usepackage{pifont}
\usepackage{colortbl}
\usepackage{xcolor}
\usepackage{hyperref}
\usepackage{amssymb} 
\definecolor{wacvblue}{rgb}{0.21,0.49,0.74}

\def\wacvPaperID{1165} 
\def\confName{WACV}
\def\confYear{2027}

\title{SEER-VAR: Semantic Egocentric Environment Reasoner \\ for Vehicle Augmented Reality}

\author{Yuzhi Lai$^1$
\and
Shenghai Yuan$^2$
\and
Peizheng Li$^{13}$
\and
Jun Lou$^3$
\and
Zhuoyu Ran$^1$ 
\and
Andreas Zell$^{1*}$ \\
$^1$ Eberhard Karls Universität Tübingen, $^2$ Nanyang Technological University, \\ $^3$ 
Mercedes-Benz AG, $^*$ Corresponding Author
\\ {\tt\small yuzhi.lai@university-tuebingen.de}
}

\begin{document}
\maketitle
\begin{abstract}
Augmented reality (AR) promises a flexible paradigm for future human-vehicle interaction. However, AR for driving scenarios poses a fundamental challenge: an egocentric camera simultaneously observes two geometrically distinct domains, \ie the vehicle cabin and the dynamic road. Existing monocular SLAM and AR systems implicitly assume a single coherent world coordinate frame, causing unstable spatial anchoring. To address this, we formulate driving AR as a dual-context egocentric perception problem and propose \textit{SEER-VAR}, a novel framework that disentangles interior and exterior observations from a monocular wearable camera and tracks them under separate reference frames. Our system performs depth-guided semantic decomposition and maintains two synchronized SLAM branches to estimate 6-DoF camera poses relative to intra- and extra-vehicle contexts. A vision-language module generates context-aware AR overlays conditioned on scene understanding and vehicle state. For evaluation, we introduce EgoSLAM-Drive, a real-world driver-egocentric dataset with dual-context 6-DoF ground truth and AR anchoring annotations across diverse driving scenarios. Experiments show that our formulation improves intra- and extra-tracking robustness, yielding spatially coherent, user-endorsed AR guidance. 
\end{abstract}
    
\section{Introduction}
\label{sec:intro}
Recent advances in embodied AI, language understanding, and egocentric vision are enabling human-centered perception and interaction systems \cite{grauman2022ego4d}. In safety-critical domains such as driving, AI-powered AR can flexibly support applications like navigation guidance, hazard highlighting, driver coaching, and in-cabin assistance by overlaying task-relevant cues in the user’s first-person view \cite{gabbard2014behind, riener2018special}. However, deploying such systems in real-world mobile environments remains challenging: the internal context (vehicle cabin) and external context (road scene) are distinct and evolve continuously, demanding robust real-time perception, localization, and semantic reasoning \cite{haselberger2024situation, 8575524}. 

\textbf{Existing} works in egocentric SLAM and AR often assume static or constrained environments \cite{8658783} and focus on either visual localization \cite{liu2022egocentric} or content rendering \cite{xu2022arshopping}. 
As a result, single-branch pipelines \cite{xu2025enhancing} often overfit to the dominant context, making it difficult to maintain stable localization and consistent AR anchoring across the in-cabin and on-road spaces \cite{mur2015orb}. This dual-view egocentric grounding enables a new class of SLAM-and-AR-driven Head-Up-Displays (HUDs) and an assistant for future highly automated vehicles (e.g., Tesla robocar w.o. steering) as fully customizable, user-centric interfaces rendered in AR. Moreover, the same capability can improve safety in today’s vehicles by presenting key information and alerts in the driver’s line of sight, reducing off-road glances and supporting more continuous attention to the road \cite{8466859}.
\begin{figure}[t]
      \centering
      \includegraphics[width=0.4\textwidth]{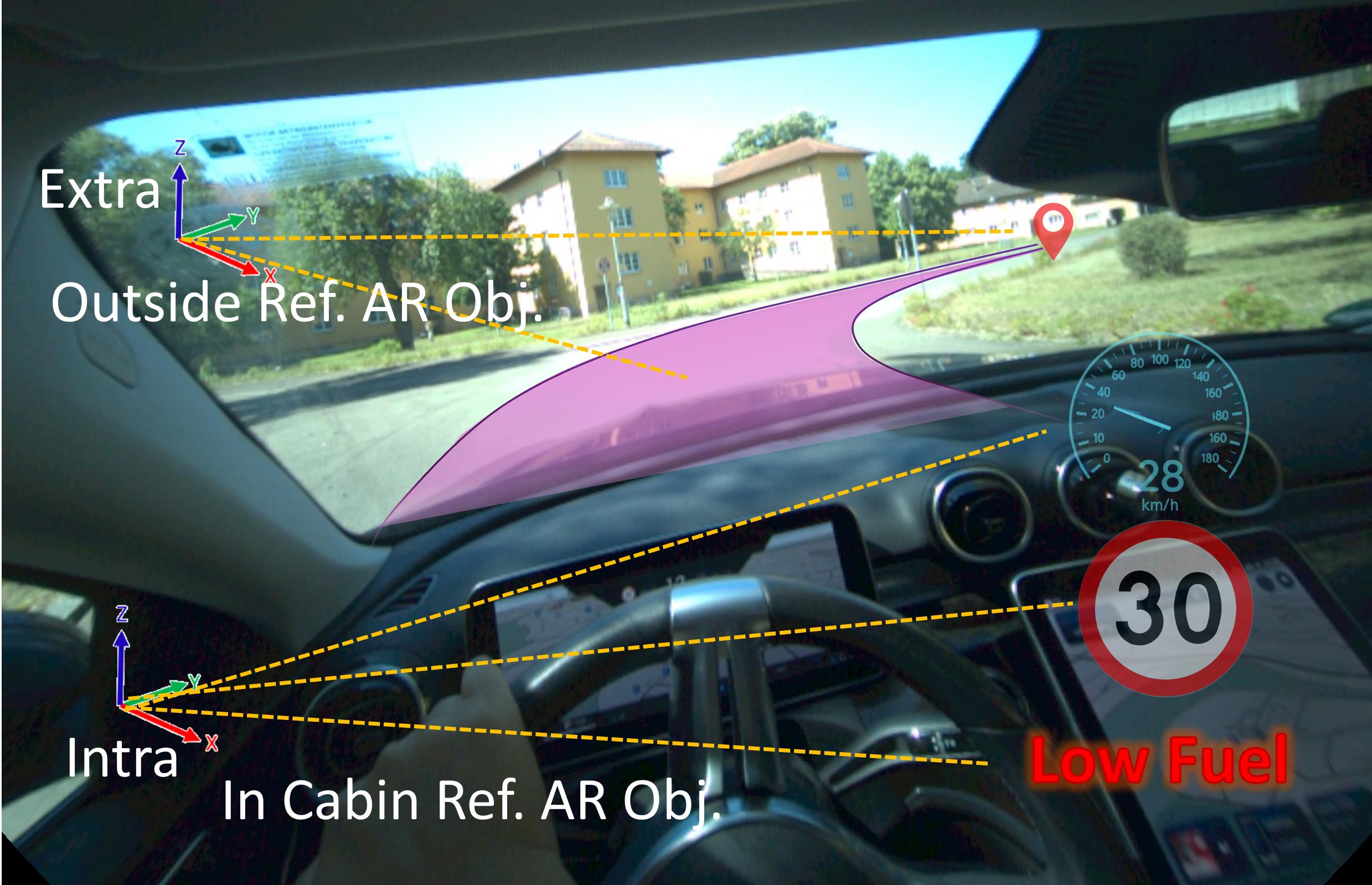}
      \caption{SEER-VAR separates the cabin and road contexts and flexibly provides LLM-driven driving assistance as needed.}
      \label{future}
      \vspace{-10pt}
\end{figure} %

The \textbf{challenges} are threefold: (1) reliable localization and mapping in spatially entangled multi-context scenes from monocular egocentric video; (2) decomposing mixed observations into semantically meaningful contexts for task-relevant perception; and (3) generating spatially grounded AR overlays that remain coherent as driving conditions and user intent evolve.

We propose \textbf{S}emantic \textbf{E}gocentric \textbf{E}nvironment \textbf{R}easoner for \textbf{V}ehicle \textbf{A}ugmented \textbf{R}eality (\textbf{SEER-VAR}), a novel framework shown in \cref{future} that combines SLAM and semantic decomposition. It decouples cabin and road scenes via separate reference frames to enable stable egocentric AR guidance in driving. SEER-VAR first decomposes raw RGB video into spatially-aware cabin and road views via data-driven context separation. Two separate SLAM branches then estimate the 6-DoF camera trajectory for each context. Building on these geometric estimates, a pretrained LLM generates context-aware AR guidance such as dashboards, navigation cues, and hazard alerts. To benchmark dual-context SLAM and AR anchoring, we also introduce EgoSLAM-Drive, a real-world driver-egocentric dataset, and evaluate SEER-VAR with user studies.

Our main contributions are summarized as follows:

\begin{itemize}
    
    
    
     \item \textbf{SEER-VAR framework.} We propose SEER-VAR, an egocentric AR framework that decouples in-cabin and on-road scenes into separate reference frames and performs dual-context pose tracking for stable AR anchoring. 

    \item \textbf{Semantic decomposition.} We introduce a depth-guided context decomposition approach that integrates vision-language grounding with a data-driven discriminator to separate egocentric observations into interior and exterior vehicle contexts, while suppressing dynamic distractors. 

    \item \textbf{Conditioned AR decision module.} We develop a GPT-based module that produces context-aware AR overlay decisions (e.g., dashboard cues, hazard alerts) conditioned on egocentric observations and vehicle state, enabling adaptive content selection and placement. 

  \item \textbf{EgoSLAM-Drive dataset and validation.} We introduce EgoSLAM-Drive, a real-world driver-egocentric dataset with dual-context 6-DoF ground truth and AR anchoring annotations, and provide a benchmark showing that existing methods struggle in this setting while SEER-VAR achieves robust tracking and improved user experience. The dataset is available on \hyperlink{blue}{https://github.com/laiyuzhi/EgoDrive}.

\end{itemize}
\section{Related Works}
\label{sec:Related_Work}

\textbf{Egocentric SLAM and Mapping.}
Egocentric SLAM systems such as ORB-SLAM3~\cite{ORBSLAM3_TRO}, SchurVINS~\cite{fan2024schurvins}, and AirSLAM~\cite{xu2025airslam} enable robust first-person localization in static environments. DROID-SLAM~\cite{teed2021droid} jointly optimizes camera poses and dense depth via recurrent updates to improve robustness under dynamics. To handle moving content, DynaSLAM~\cite{8421015} and MaskSLAM~\cite{8575524} remove dynamic regions via segmentation, while CTO-SLAM~\cite{li2024cto} and MaskFusion~\cite{8613746} pursue object-level modeling and tracking and demonstrate improved handling of dynamic or articulated elements. However, these pipelines still largely assume a single coherent background scene reference, and thus do not address dual-context egocentric views where cabin and road background scenes require separate reference frames for stable anchoring. On the data side, Ego4D~\cite{grauman2022ego4d} and EPIC-Kitchens~\cite{damen2022rescaling} focus on egocentric activity understanding, while DR(eye)VE~\cite{7789504} and GEM~\cite{akbiyik2023routeformer} target driver intention prediction. None of these datasets target dual-context localization in driving.

\noindent \textbf{Vision-Language Grounding.}
Recent vision-language models enable open-set detection and zero-shot segmentation via natural language prompts. Grounding DINO~\cite{GroundingDINO} and Grounded SAM~\cite{GroundedSAM}, built on Segment Anything~\cite{kirillov2023sam}, show strong generalization and are widely used for offline annotation. Their computational cost, however, makes naive per-frame use impractical for real-time egocentric AR. More broadly, common grounding pipelines provide limited mechanisms to exploit spatiotemporal semantic structure for consistency and compute-aware scheduling, which constrains stable, geometry-aware AR overlays in mobile settings.

\begin{figure*}[t]
      \centering
      \includegraphics[width=0.80\textwidth]{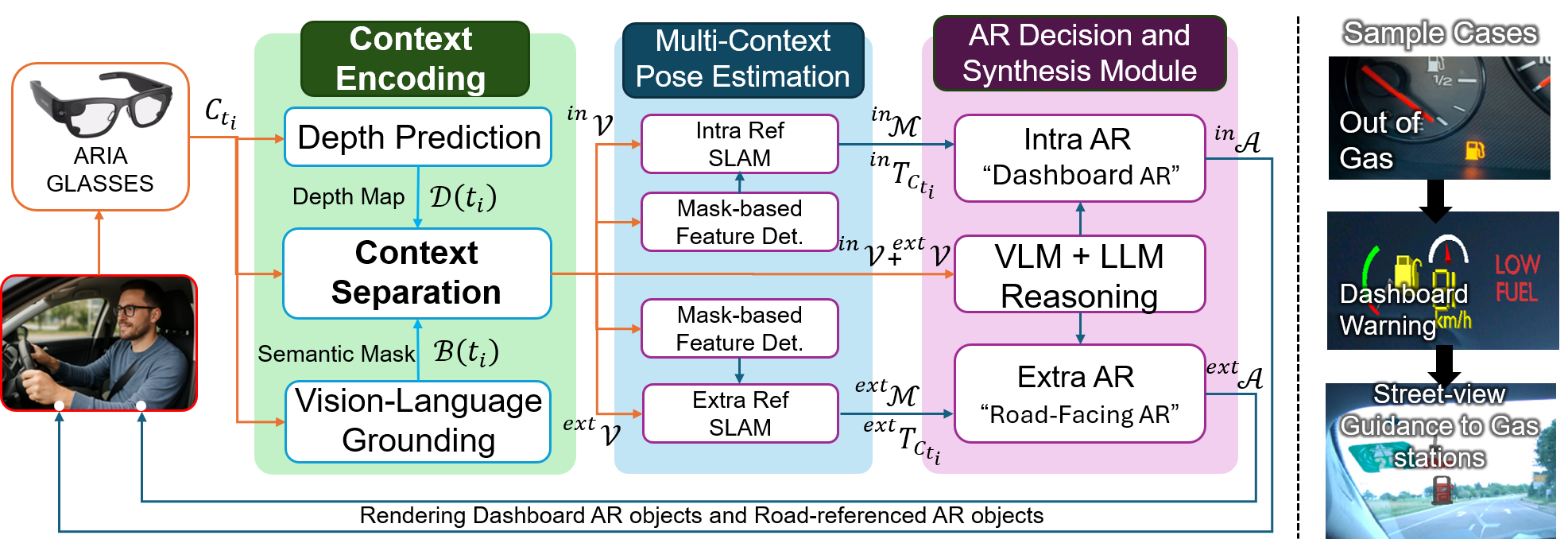}
    
      \caption{Our system consists of three modules: context encoding, multi-context pose estimation, and AR decision and synthesis module.} 
      \label{overview} 
   \vspace{-10pt}
\end{figure*} %

\noindent \textbf{Language Agents for Augmented Reality.}
Large language models (LLMs) such as GPT-4~\cite{openai2023gpt4} have demonstrated strong abilities in instruction following~\cite{brohan2023can}, multimodal reasoning~\cite{alayrac2022flamingo, shi2026correct}, and task planning~\cite{11534298}. Recent work has begun integrating LLMs into embodied agents~\cite{shridhar2022cliport, xu2022seer, 2025exploringcontextawarellmdrivenlocomotion, Li_2026_CVPR, wang2026cruise}, enabling language-driven decision-making. Systems like IM~\cite{huang2022innerMonologue} show that prompting LLMs can guide task decomposition in structured settings. Recent work  \cite{2025exploringcontextawarellmdrivenlocomotion} demonstrates a context-aware LLM-driven navigation system for VR. However, most approaches are limited to static environments and overlook spatially grounded data such as SLAM outputs. While SituationAdapt \cite{li2024situationadapt} employs LLMs to adjust UI layout at the semantic level to minimize occlusion of salient objects in the scene, they still fail to account for the existence of multiple spatial contexts—such as distinct coordinate frames for in-vehicle and out-of-vehicle environments—surrounding the user. This hinders their robustness in dynamic, blended indoor-outdoor scenarios. In particular, the use of LLMs for generating contextual, real-time overlays in egocentric AR remains underexplored.

\noindent \textbf{AR Systems.}
Traditional AR often relies on static scenes and predefined content, with limited support for robust spatial anchoring in the wild \cite{lai2026objectselectionmarkerlessgazebasedrobot}. Recent systems such as ARShopping \cite{xu2022arshopping} and ARShoe \cite{an2021arshoe, shi2025fashionpose} demonstrate the feasibility of real-time AR recommendations on consumer devices, while Vicarious \cite{zaman2023vicarious} explores context-aware viewpoint selection in mixed reality. In the automotive domain, HUDs aim to keep critical information within the driver’s line of sight, helping reduce off-road glances and increase safety \cite{4538831}. However, conventional HUDs are constrained by a small display area and thus can present limited content \cite{8466859}. In contrast, AR-HUD can reduce off-road glances and workload \cite{narzt2006augmented}. AR-HUD often fall into two paradigms: screen-fixed overlays that render UI at a constant screen location \cite{gabbard2014behind, 8466859}, and world-anchored overlays that align content with scene elements via detection/tracking \cite{pomarjanschi2012gaze, jansen2026mirage}. Screen-fixed overlays lack geometric grounding and semantic understanding, while world-anchored overlays depend on object detection and can degrade when the target region is occluded or tracking becomes unstable. 

\section{Problem Formulation}

Our system addresses the problem of dual-context egocentric AR in driving. In this scenario, the input consists of an RGB video stream $\{\mathcal{I}(t_i)\}_{i=1}^{n} \in \mathbb{R}^{3 \times H \times W}$ (1408 × 1408), capturing both the static intra-vehicle scene and the dynamic extra-vehicle scene.
The core goal is to accurately distinguish and model these two environmental components to enable context-aware egocentric AR recommendation
, such as dashboard widgets or hazard warnings inside the cabin ${}^{in}\mathcal{A}$, and context-specific visual enhancements such as navigation hints in the external environment ${}^{ext}\mathcal{A}$. Unlike conventional egocentric SLAM/AR formulations that assume a single world frame, this requires distinguishing and modeling the two contexts under separate reference frames to avoid drifting or context-confused anchoring.
 


We define the reference frames
used in this paper. The \textbf{intra frame} $\prescript{in}{}{(\cdot)}$ serves as a fixed coordinate in the vehicle cabin, while the  \textbf{extra frame} $\prescript{ext}{} {(\cdot)}$ describes a fixed coordinate system in the external environment. The \textbf{camera frame} $\prescript{{C}_{t_i}}{}{(\cdot)}$ defines the coordinate system for the RGB camera on ARIA glasses. The transformation $T \in \mathbb{SE}(3)$ denotes a rigid body transformation in the special Euclidean group. The ARIA glasses worn by the driver act as a moving camera $C_{t_i}$. At time $t_i$, the pose of the camera relative to the two frames is represented by ${}^{in}T_{C_{t_i}}$ and ${}^{ext}T_{C_{t_i}}$, where 
${}^{in}T_{C_{t_i}}$ maps points from the camera frame to the intra frame, and ${}^{ext}T_{C_{t_i}}$ maps points to the extra frame. All notations used in our paper are provided in Supp. Tab. 9.

\section{Methodology}

We present an egocentric AR recommendation system, driven by context-aware SLAM branches combined with GPT-based semantic recommendation. Our approach simultaneously tracks user motion and semantic environment information in intra- and extra frames, enabling context-aware recommendations and precise anchoring of AR overlays. The overview of our system is shown in \cref{overview}.

\subsection{Context Encoding}

In our proposed approach, 
we first utilize Depth Anything V2 \cite{yang2024depth} to    generate the depth map $\{\mathcal{D}(t_i)\}_{i=1}^{n} \in \mathbb{R}^{H\times W}$, encoding geometric information about the scene. We then identify potential dynamic objects from $\{\mathcal{I}(t_i)\}_{i=1}^{n}$, leveraging a combination of visual-language grounding and semantic segmentation. We utilize YOLOE26~\cite{yolo26e_ultralytics} to obtain the bounding boxes of the objects. Specifically, we first define two sets of representative semantic objects: $\{ {c_{in}(t_i)} \}_{i=1}^{n}$ for intra-vehicle context (e.g., dashboard, steering wheel, seats, hands) and $\{ {c_{ext}(t_i)} \}_{i=1}^{n}$ for extra-vehicle context (e.g., road, vehicles, pedestrians, traffic signs, buildings). From these detected objects, we further define a subset of potential dynamic objects $c^{dyn}$.
     EdgeSAM \cite{zhou2025edgesam} generates the semantic masks for potential dynamic objects $\{\mathcal{B}(t_i)\}_{i=1}^{n} \in \mathbb{R}^{H\times W}$. 

\begin{figure}[t]
      \centering
      
      \includegraphics[width=0.43\textwidth]{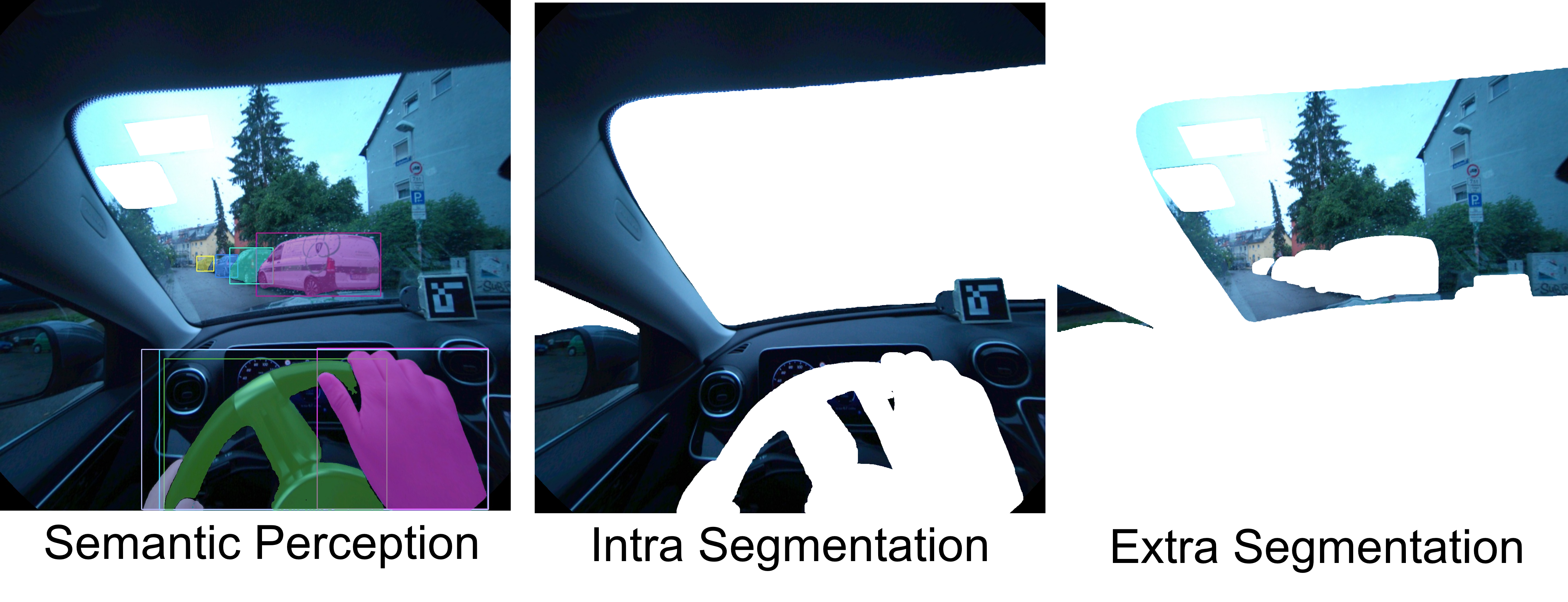}
    \vspace{-5pt}
      \caption{
      Example results of intra-extra Segmentation.} 
      \label{resultseg}
     \vspace{-10pt}
\end{figure}

\begin{algorithm}[t]
\scriptsize
\caption{Data Driven Discriminator}
\label{alg:filter}
\textbf{Input}: RGB $\mathcal{I}(t_i)$, depth map $\mathcal{D}(t_i)$, mask $\mathcal{B}(t_i)$, object sets $c_{in}(t_i)$, $c_{ext}(t_i)$\\
\textbf{Output}: Intra-extra vehicle images $\mathcal{V}_{in}(t_i)$, $\mathcal{V}_{out}(t_i)$
\begin{algorithmic}[1]
\If{$c_{in}(t_i)=\emptyset$} \Comment{Context Check} 
\State Compose RGBA images \quad ${}^{ext}\mathcal{V}(t_i)$
\State \textbf{return} ${}^{ext}\mathcal{V}(t_i)$
\ElsIf{$c_{ext}(t_i)=\emptyset$}
\State Compose RGBA images \quad ${}^{in}\mathcal{V}(t_i)$
\State \textbf{return} ${}^{in}\mathcal{V}(t_i)$
\EndIf
\State Data Augmentation: $H(k)$ over $\mathcal{D}(t_i)$: \Comment{Depth-based Separation} \\$H(k)\gets \# \left \{ (x, y) \mid \mathcal{D}(x, y)=k  \right \}, k = 0, 1,2, ..., 255 $
\State Find first peak: $k_1 \gets \underset{k}{argmax}H(k)$
\State Compute second peak: $k_2 \gets \underset{k}{argmax}(k-k_1)^{2}\cdot H(k)$ 
\State Get threshold: $k_{min}\gets \underset{k\in\left [ min(k_1, k_2), max(k_1, k_2) \right ] }{argmin}H(k)$
\State Mask out semantic dynamic object and generate binary intra mask for:  ${}^{in}\mathcal{V}$: ${}^{in}\mathcal{V}^{\alpha}(t_i) \gets ( 1 - \mathcal{B}(t_i))\odot \mathds{1}(\mathcal{D}(t_i) < k_{min})$
\State Mask out semantic dynamic object and generate binary extra mask for ${}^{ext}\mathcal{V}(t_i)$: ${}^{ext}\mathcal{V}^{\alpha}_{ext}(t_i) \gets ( 1 - \mathcal{B}(t_i))\odot \mathds{1}(\mathcal{D}(t_i) > k_{min})$ 
\State Compose RGBA images \quad ${}^{in}\mathcal{V}(t_i),{}^{ext}\mathcal{V}(t_i)$
\State \textbf{return} ${}^{in}\mathcal{V}(t_i),{}^{ext}\mathcal{V}(t_i)$
\end{algorithmic}
\end{algorithm}
Subsequently, we separate intra- and extra-vehicle regions using a data-driven depth discriminator, as shown in \cref{alg:filter}. The discriminator is divided into two steps: context checking and depth-based separation. We first check the representative context objects detected in the current frame. If the frame is dominated only by the set $c_{in}$ or $c_{ext}$, it is treated as a single-context observation and assigned directly to the corresponding branch. Depth-based separation is applied only when both intra- and extra-vehicle contexts are present in the same egocentric view.
The key observation is that, in driving egocentric views, pixels belonging to the cabin and the outside scene typically form a bimodal depth distribution: near depths correspond to interior structures (e.g., dashboard) while far depths correspond to the road environment.
To exploit this property robustly, we compute a 256-bin depth histogram $H(k)$ from the depth map $\mathcal{D}(t_i)$. We first estimate a per-frame separation threshold $k_{min}$ as the valley between the two dominant modes. To avoid bias from transient moving objects, we then exclude pixels in the dynamic mask $\mathcal{B}(t_i)$. We then obtain the two binary masks ${}^{in}\mathcal{V}^{\alpha}(t_i) \in \mathbb{R}^{H \times W}$ and ${}^{ext}\mathcal{V}^{\alpha}(t_i) \in \mathbb{R}^{H \times W}$ by thresholding the depth map. Finally, we concatenate each mask with the RGB frame $\mathcal{I}(t_i)$ to form two RGBA streams $ {}^{in}\mathcal{V}(t_i)$ and $ {}^{ext}\mathcal{V}(t_i) \in \mathbb{R}^{4 \times H \times W}$.
Qualitative examples are shown in \cref{resultseg}. This adaptive thresholding avoids using a fixed depth cut, which is brittle under changing scale and illumination. For reproducibility and examples in different situations, see Supp. Sec. 11 and 21.

\subsection{Multi-Context Pose Estimation}

To support spatially stable AR anchoring in driving, we propose Context-Aware SLAM Branches (CASB), which estimate camera motion under two reference frames corresponding to the in-cabin and on-road contexts. The key challenge is not only dynamic objects, but context competition: intra features (including hands, dashboard, etc.) and extra features (including traffic, landmarks, etc.) co-exist in the same view and can dominate at different times. A single-branch SLAM is thus forced to explain two geometrically distinct domains with one pose track, often drifting toward the dominant context and yielding inconsistent anchoring across the other. Conventional feature-based SLAM (e.g., ORB-SLAM \cite{mur2015orb}) relies on RANSAC \cite{fischler1981random} to reject outliers, which becomes brittle when a large fraction of correspondences comes from context-irrelevant regions.


CASB addresses this by estimating two context-conditioned camera trajectories w.r.t. two reference frames: we restrict feature matching to each context using the binary masks produced by the context encoding module, and compute correspondences independently for the two domains,
${}^{in}\mathcal{F}(t_i) = \text{SuperPoint}(\mathcal{I}(t_i) \odot \mathcal{}^{in}{V}^{\alpha}(t_i)) $,
where $\odot$ denotes element-wise masking and $\text{SuperPoint}$ \cite{detone2018superpoint} is used to compute sparse and robust feature correspondences under rapid head motion (The backend follows ORB-SLAM3 \cite{ORBSLAM3_TRO}). The restricted features ${}^{ext}\mathcal{F}(t_i)$ of the extra-frame are calculated in the same way. ${}^{in}\mathcal{F}(t_i)$ and ${}^{ext}\mathcal{F}(t_i)$ are then passed to two separate SLAM branches of CASB, producing camera poses ${}^{in}T_{C_{t_i}}$, ${}^{ext}T_{C_{t_i}}$ and spatial maps ${}^{in}\mathcal{M}, {}^{ext}\mathcal{M} \in \mathbb{R}^{n\times 3}$ for both frames.


\subsection{GPT-based Semantic Recommendation}

To enable semantic-aware AR overlay generation, we introduce a GPT-based recommendation system that integrates structured scene perception with vehicle status for semantic reasoning and spatial content generation.

The system processes ${}^{in}\mathcal{V}(t_i)$ and ${}^{ext}\mathcal{V}(t_i)$ along with the structured vehicle state $\mathcal{X}(t_i)$ (e.g., fuel level, driving time). It is invoked only at event triggers (e.g., low-fuel warnings, SLAM initialization, or explicit user requests) to avoid unnecessary latency.
At each trigger, the agent $\pi$ is prompted with a structured checklist that summarizes the current context:
(1) What type of environment is the vehicle in? (2) What is the current status of the vehicle and driver (e.g., dashboard visibility, fuel level)? (3) What type of AR content is relevant in the current situation? (4) Where should the covering layer be anchored in the environment while ensuring rich features? It is constrained to select (i) the relevant overlay category and (ii) an image-space anchoring box in the corresponding context. Formally, 

\begin{align}
 \pi : {}^{in}\mathcal{V}(t_i), {}^{ext}\mathcal{V}(t_i), \mathcal{X}(t_i) \mapsto  {}^{in}\mathcal{L}\times {}^{in}b, {}^{ext}\mathcal{L}\times {}^{ext}b
\end{align}
where ${}^{in}\mathcal{L}, {}^{ext}\mathcal{L}$ are the labels of the recommended overlay (e.g., dashboard, navigation hint, service advertisement, parking information  ). ${}^{in}b, {}^{ext}b \in \mathbb{R}^4$ are the predicted bounding boxes in image space that specify the recommended anchoring region for overlays.



\begin{figure}[t]
      \centering
      \includegraphics[width=0.4\textwidth]{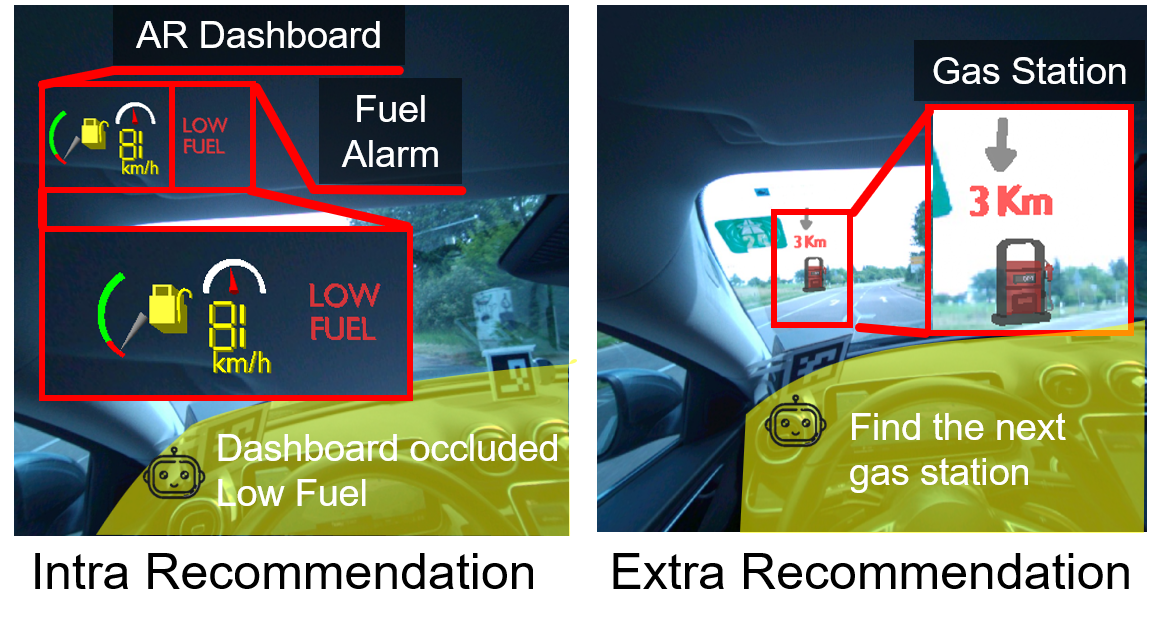}
      \vspace{-8pt}
  
      \caption{
      Example of GPT-based semantic recommendation. }
      \label{ar}
 \vspace{-10pt}
\end{figure}

As shown in  \cref{ar} (left), when the physical dashboard is occluded and fuel is low, our recommendation system suggests two intra-vehicle overlays: a virtual dashboard and a fuel alarm. As shown in  \cref{ar} (right), the system identifies a relevant external navigation hint (i.e., a gas station) and determines an appropriate 3D anchoring pose on the map for overlay rendering, enabling globally consistent guidance. Examples are provided in Supp. Sec. 9. Detailed prompt engineering is provided in Supp. Sec. 15.



\subsection{AR Rendering}

To provide immersive and context-aware experiences for drivers, our system renders an AR overlay in both intra- and extra-frame. 
Given the recommendation bounding box ${}^{in}b(t_i)$ from the policy $\pi$ and the results of CASB ${}^{in}T_{C_{t_i}}$, the system renders digital overlays anchored inside the vehicle cabin. To project the AR content stably within the cabin, we estimate a 3D anchor point ${}^{in}P_{\mathcal{A}_{in}} \in \mathbb{R}^3$ in the intra frame using a predefined fixed virtual depth $z_{a}$. Specifically, let the center of the bounding box ${}^{in}b(t_i)$ be $({}^{in}u, {}^{in}v)$. Then the anchor point in the intra frame is calculated by back-projecting this 2D point with depth $z_{a}$ and the intrinsic matrix of the ARIA glasses RGB camera $\mathcal{K}\in\mathbb{R}^{3\times3}$:

\begin{align}
{}^{in}P_{\mathcal{A}_{in}} = \left [ x_a,y_a,z_a \right ]^T = z_{a}\cdot\mathcal{K}^{-1}\begin{bmatrix}
{}^{in}u, {}^{in}v, 1
\end{bmatrix}^T
    \label{backpro}
\end{align}

Since rendering is done from the camera’s perspective, the relative 3D rendering position ${}^{C_{t_i}}P_{\mathcal{A}_{in}}$ is the following:

\begin{align}
{}^{C_{t_i}}P_{\mathcal{A}_{in}} = ({}^{in}T_{C_{t_i}})^{-1} \cdot {}^{in}P_{\mathcal{A}_{in}}
\label{transformrender}
\end{align}

For rendering external content, the bounding box $b_{ext}(t_i)$ is projected onto the map $\mathcal{M}_{ext}$. We collect all 3D map points inside the bounding box and compute the anchor point ${}^{ext}P_{\mathcal{A}_{ext}}$ as the center of those projected map points. To render the overlay from the current camera view, we compute the  relative 3D rendering position ${}^{C_{t_i}}P_{\mathcal{A}_{ext}}$ with:

\begin{align}
{}^{C_{t_i}}P_{\mathcal{A}_{ext}} = ({}^{ext}T_{C_{t_i}})^{-1} \cdot {}^{ext}P_{\mathcal{A}_{ext}}
\label{transformrender2}
\end{align}

In addition to anchoring the overlay position, we also enforce orientation consistency using the rotation component of the estimated poses. Specifically, each pose ${}^{in}T_{C_{t_i}}$ and ${}^{ext}T_{C_{t_i}}$ is decomposed to get the orientation ${}^{in}R_{C_{t_i}}$ and ${}^{ext}R_{C_{t_i}}$. During rendering, we align the overlay’s local orientation with the corresponding orientation, so that the AR content remains front-facing and stable under head rotations. Intuitively, as the user turns their head, the overlay orientation is updated consistently with the head rotation, ensuring that the AR overlays stay properly oriented in the corresponding intra- or extra-vehicle reference frame.

As shown in Supp. Sec. 12, our rendering pipeline achieves over 100 FPS. We ensure safety through semi-transparent, non-intrusive overlays. In the meantime, the AR content is optional and can be disabled entirely when not needed. Details are discussed in Supp. Sec.18. Robustness analysis of anchoring is presented in Supp. Sec. 14. 


\section{EgoSLAM-Drive Dataset}
 \begin{figure}[th]
      \centering
  
      \includegraphics[width=0.4   \textwidth]{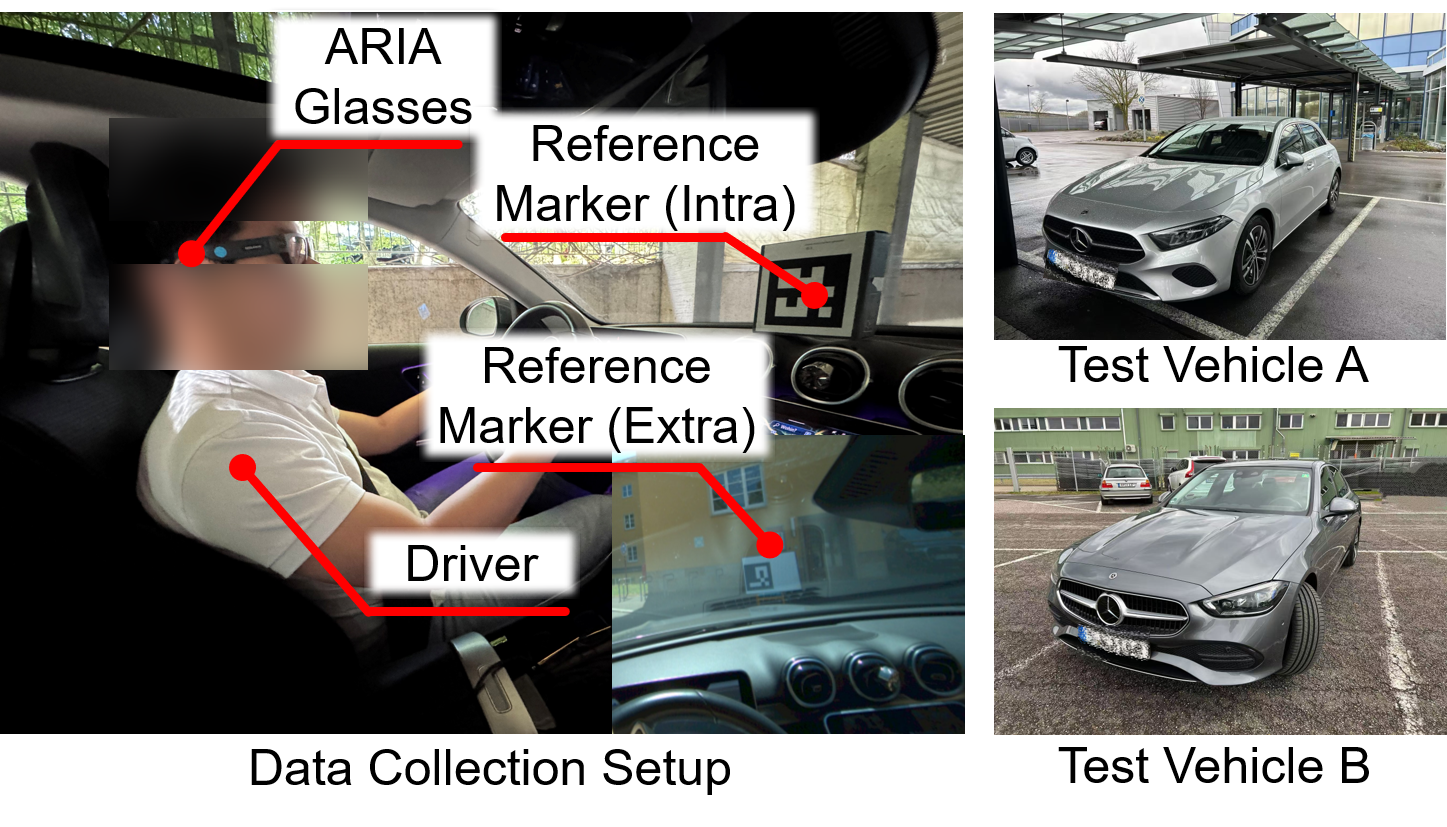}
      \vspace{-8pt}
     
      \caption{ Driver and vehicles setup
   }
      \label{collect}
   
\end{figure} %
\begin{figure*}[th]
      \centering
      \includegraphics[width=0.85\textwidth]{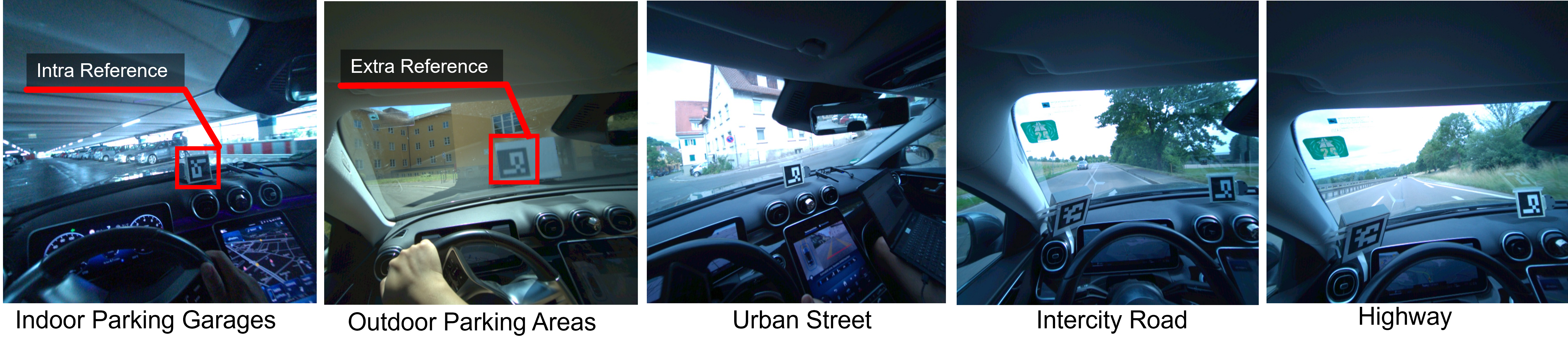}
      \vspace{-12pt}
      \caption{
      Example from the EgoSLAM-Drive dataset. This figure demonstrates the diversity in our dataset.} 
      \label{datasetup}
     \vspace{-10pt}
\end{figure*} %
\begin{table*}[t]
     \centering
 \footnotesize
\setlength{\tabcolsep}{5pt} 
 \caption{Comparison of existing datasets.}
  \label{tab:databaseline}
 \begin{tabular}{l|c|c|cc|c|c}
 \toprule
 \multirow{3}{*}{Dataset}  &   \multirow{3}{*}{Sensors}  & \multirow{3}{*}{\shortstack{Driver\\Egocentric?}}& \multicolumn{2}{c|}{Challenges} & \multirow{3}{*}{\shortstack{Multi-Context \\ GT?}}& \multirow{3}{*}{\shortstack{Purpose}}\\
  & & & \multirow{2}{*}{\shortstack{Dynamic \\ Object?}}& \multirow{2}{*}{\shortstack{Dual \\ Context?}}&& \\ 
&&&&&  \\
\midrule
  KITTI \cite{Geiger2013IJRR} & RGB/Grey, IMU, GNSS  & \textcolor{red}{\ding{55}}  & \textcolor{green}{\ding{51}}  & \textcolor{red}{\ding{55}} & \textcolor{red}{\ding{55}}& SLAM \\
  TartanAir \cite{9341801}  &  RGB-D, LiDAR   & \textcolor{red}{\ding{55}}   & \textcolor{green}{\ding{51}} & \textcolor{red}{\ding{55}}  & \textcolor{red}{\ding{55}}& SLAM \\
 TUM RGB-D \cite{sturm2012benchmark} & RGB-D &  \textcolor{red}{\ding{55}} & \textcolor{green}{\ding{51}} & \textcolor{red}{\ding{55}}  & \textcolor{red}{\ding{55}}& SLAM \\
4Seasons \cite{wenzel20204seasons} & RGB, IMU, GNSS  &  \textcolor{red}{\ding{55}} &\textcolor{green}{\ding{51}} &  \textcolor{red}{\ding{55}}  & \textcolor{red}{\ding{55}}& SLAM \\
 VBR \cite{brizi2024vbr} &  RGB, LiDAR, IMU, GNSS  & \textcolor{red}{\ding{55}}  &\textcolor{gray}{\ding{51}\rotatebox[origin=c]{-9.2}{\kern-0.60em\ding{55}}} & \textcolor{red}{\ding{55}}  & \textcolor{red}{\ding{55}} & SLAM\\
 Hilti-Oxford \cite{zhang2022hilti}  & RGB, LiDAR   & \textcolor{red}{\ding{55}}  &  \textcolor{red}{\ding{55}}  &  \textcolor{red}{\ding{55}}   &  \textcolor{red}{\ding{55}} & SLAM \\
 nuScenes \cite{caesar2020nuscenes}   & RGB, LiDAR, IMU, GNSS  & \textcolor{red}{\ding{55}}  & \textcolor{green}{\ding{51}} & \textcolor{red}{\ding{55}}  &\textcolor{red}{\ding{55}} & SLAM\\
 Waymo \cite{sun2020scalability}  & RGB, LiDAR  & \textcolor{red}{\ding{55}} & \textcolor{green}{\ding{51}} &  \textcolor{red}{\ding{55}} &\textcolor{red}{\ding{55}} & SLAM\\
 AEA \cite{lv2024aria}  & RGB, IMU, GNSS & \textcolor{red}{\ding{55}}  &\textcolor{green}{\ding{51}}  &  \textcolor{red}{\ding{55}} & \textcolor{red}{\ding{55}}& SLAM \\
 A2PD \cite{kong2025aria} & RGB, IMU, GNSS & \textcolor{red}{\ding{55}} &\textcolor{green}{\ding{51}}  &  \textcolor{red}{\ding{55}} & \textcolor{red}{\ding{55}} & SLAM\\
 LaMAria \cite{krishnan2025benchmarking}   & RGB, IMU, GNSS  & \textcolor{red}{\ding{55}} & \textcolor{green}{\ding{51}} & \textcolor{gray}{\ding{51}\rotatebox[origin=c]{-9.2}{\kern-0.60em\ding{55}}} &  \textcolor{red}{\ding{55}}& SLAM  \\
  DR(eye)VE  &RGB, Eye Tracker&\textcolor{green}{\ding{51}}&\textcolor{green}{\ding{51}}&\textcolor{green}{\ding{51}}&\textcolor{red}{\ding{55}}&Intent Recognition\\
 \midrule
\textbf{Ours} &RGB, IMU, GNSS  & \textcolor{green}{\ding{51}} & \textcolor{green}{\ding{51}} & \textcolor{green}{\ding{51}}  & \textcolor{green}{\ding{51}} & SLAM \\
 \bottomrule
 \end{tabular}
 \vspace{-5pt}
\end{table*}

\noindent\textbf{Datasets and benchmarks:} As shown in \cref{tab:databaseline}, there exist    many well-established SLAM datasets. However, most do not reflect the motion patterns and viewpoint changes of drivers in real-world scenarios. 
EgoSLAM-Drive targets a distinct setting: dual-context egocentric driving, where in-cabin and on-road scenes are simultaneously observed and spatially entangled. This introduces two key challenges: (1) multi-context localization (cabin vs. road), and (2) heavy dynamic interference (e.g., hands inside the cabin~\cite{shi2026sgmhand}, pedestrians and vehicles outside). EgoSLAM-Drive provides 6-DoF ground truth for both intra- and extra-vehicle reference frames, enabling benchmarking of multi-reference egocentric SLAM and AR anchoring. 

\noindent\textbf{Device and sensors:} We collect all data using Meta ARIA glasses, including an RGB camera (1408×1408, 20 FPS, 110 FOV) and two factory-calibrated inertial measurement units (IMUs) operating at 1 kHz and 800 Hz, respectively. All sensor streams are accurately timestamped and mounted on a rigid frame, ensuring consistent spatial alignment. To improve generalization, we recorded our dataset with two vehicles. Due to the electromagnetic shielding effect of the vehicle carriage, the built-in GNSS of the Aria glasses is unavailable. Therefore, we use an external GNSS receiver to obtain the vehicle's trajectory. 

\noindent\textbf{Environments and setup:} As shown in \cref{datasetup}, our dataset was collected while the driver wore ARIA glasses during real-world driving with diverse weather and lighting conditions, including indoor parking garages, outdoor parking areas, urban streets, intercity roads, and highways.
Following common practice in SLAM benchmarking \cite{7989443, wang2026seeing}, we use ArUco markers as sparse 6-DoF ground-truth references. For 6-DoF motion ground truth of intra-frame, markers are placed inside the cabin and used whenever visible. For extra-frame ground truth, we place 80 cm ArUco boards approximately every 5 m along the roadside in safe environments (urban streets and outdoor parking area).  To ensure label accuracy, an external marker is considered valid only when its image footprint exceeds 100 pixels, reducing ambiguity from long-range detections. In total, EgoSLAM-Drive contains 71 sequences ranging from 1015 to 7338 frames ($>$80 minutes in total).  For details (e.g. speed, weather) and visualizations, see Supp. Sec. 8 and 17.


\section{Experiments}

\begin{table*}[t]
\centering
 \footnotesize
\setlength{\tabcolsep}{1pt} 
\caption{Comparison of baselines on the EgoSLAM-Drive dataset. 
“Intra Tracked?” and “Extra Tracked?” use \textcolor{green}{\ding{51}} / \textcolor{red}{\ding{55}} to indicate whether the method maintains a geometrically consistent pose in the dual contexts, respectively. 
RE, ARE, and ATE are reported as RMSE values; “--” denotes that tracking in that context is unavailable or unreliable, so metrics cannot be computed.  “Opt.” denotes the optimization process.}

\label{tab:baseline1}
\begin{tabular}{lccccccccc}
\hline
\multirow{2}{*}{Methods}  & \multirow{2}{*}{Paradigm}&
 \multirow{2}{*}{\shortstack{Intra\\Tracked?} }  &
 \multirow{2}{*}{\shortstack{Extra\\Tracked?} } &  
 \multirow{2}{*}{\shortstack{RE\\Intra (px) $\downarrow$} }  &
 \multirow{2}{*}{\shortstack{RE\\Extra (px)$\downarrow$} }&
 \multirow{2}{*}{\shortstack{ARE\\Intra ($^{\circ} $)$\downarrow$} }&
 \multirow{2}{*}{\shortstack{ARE\\Extra ($^{\circ} $)$\downarrow$} } &
 \multirow{2}{*}{\shortstack{ATE\\Intra (m)$\downarrow$} }&
 \multirow{2}{*}{\shortstack{ATE\\Extra (m)$\downarrow$} }\\ \\
\hline
MASt3R-SLAM\cite{murai2025mast3r}& Learned Feature$+$Opt.    & \textcolor{red}{\ding{55}}     & \textcolor{red}{\ding{55}}     & --   & --  & --   & -- & --   & --  \\
ORB-SLAM3 \cite{ORBSLAM3_TRO}  & ORB Feature$+$Opt.    &\textcolor{green}{\ding{51}}  & \textcolor{red}{\ding{55}}     & 
32.75 & --  & 6.54 & -- & 0.191 & --  \\
Dyna-SLAM \cite{8421015}  & Mask$+$ORB Feature$+$Opt.           &\textcolor{green}{\ding{51}} & \textcolor{red}{\ding{55}}     & 22.81 & --& 3.24 & --& 0.079 & --   \\
AirSLAM \cite{xu2025airslam} & Learned Feature$+$Opt.              & \textcolor{red}{\ding{55}}     & \textcolor{red}{\ding{55}}     & --   & --  & --   & --& --   & -- \\
sqrtVINS \cite{Peng2025TRO} & Optical Flow$+$sqrt Filter             & \textcolor{red}{\ding{55}}     & \textcolor{green}{\ding{51}}     & --   & 85.23 & --   & 48.33& --   & 2.184  \\
MaskFusion \cite{8613746}& Mask$+$ORB Feature+Opt. &\textcolor{green}{\ding{51}} & \textcolor{red}{\ding{55}}     & 23.68 & --& 3.19 & --& 0.154& -- \\

\hline
Context+ORB-SLAM3 & Dual Branch&\textcolor{green}{\ding{51}} & \textcolor{green}{\ding{51}}     & 19.04 & 26.93  & 2.46 & 6.29  & 0.086 & 0.446  \\
Context+AirSLAM & Dual Branch & \textcolor{red}{\ding{55}}& \textcolor{green}{\ding{51}} & -- & \textbf{15.38}& -- &\textbf{5.17}& -- & \textbf{0.285}\\
Context+sqrtVINS & Dual Branch & \textcolor{red}{\ding{55}} & \textcolor{green}{\ding{51}} & -- & 18.76& -- & 5.92& -- & 0.397\\
Context+MASt3R & Dual Branch & \textcolor{green}{\ding{51}}  & \textcolor{red}{\ding{55}} & 16.37 & --& 2.13& --& 0.065& --\\

\hline
\multirow{1}{*}{\shortstack{World-Anchored HUD \cite{jansen2026mirage}}} & Mask$+$AR & \textcolor{gray}{\ding{51}\rotatebox[origin=c]{-9.2}{\kern-0.60em\ding{55}}}& \textcolor{gray}{\ding{51}\rotatebox[origin=c]{-9.2}{\kern-0.60em\ding{55}}} & --&--&--&--&--&-- \\
\multirow{1}{*}{\shortstack{Screen-Fxed HUD \cite{gabbard2014behind}} } & Fixed AR& \textcolor{red}{\ding{55}}&  \textcolor{red}{\ding{55}} & --&--&--&--&--&-- \\
\hline
\textbf{Ours}      & Learned Feature$+$Dual Branch                 &\textcolor{green}{\ding{51}} &\textcolor{green}{\ding{51}} &\textbf{13.74} & {17.63}& \textbf{1.85}& {5.64}& \textbf{0.054}& {0.312} \\
\hline
\end{tabular}
\vspace{-5pt}
\end{table*}

To our knowledge, SEER-VAR is the first system to explore LLM-driven egocentric AR in vehicle settings, and since there is no directly comparable system-level baseline, we complement geometric evaluation with human user studies. 
Excluding the GPT-based recommendation module, SEER-VAR runs at 20 FPS on an NVIDIA Jetson Orin NX. All vision components are optimized with TensorRT (FP16), and the selected backbones are chosen to support deployment on edge devices. To ensure real-time capability, the GPT-based recommendation agent is implemented using the \texttt{Realtime-2.1} \cite{openai_realtime_api} with latency $<$ 1s.

\noindent\textbf{Evaluation Metrics.}
To evaluate the spatial consistency and trajectory accuracy of the estimated camera poses, we report three metrics: reprojection error (RE), absolute rotation error (ARE), and absolute trajectory error (ATE). RE measures the pixel Euclidean distance between the projected 3D ArUco landmark points and their ground-truth image detections. Detailed calculations are provided in Supp. Sec. 10. All metrics are reported as RMSE values. 

\subsection{Baseline Comparison}
\label{Benchmarking}

We compare SEER-VAR with representative SLAM, visual-inertial, dual-branch, and AR-HUD baselines on EgoSLAM-Drive. DROID-SLAM is too heavy for our system (min. 11 GB VRAM), so it is not being considered. As shown in \cref{tab:baseline1}, single-branch SLAM systems fail to maintain geometrically consistent tracking in both contexts. ORB-SLAM3, Dyna-SLAM, and MaskFusion can only track the intra-vehicle context. Conversely, sqrtVINS tracks the extra-vehicle context, but fails in the intra-vehicle context. MASt3R-SLAM and AirSLAM fail in both contexts under rapid egocentric motion, illumination changes, and dynamic traffic. A key factor is context awareness: traditional SLAM methods often overfit to dominant visual cues in either the intra-vehicle or extra-vehicle context, limiting robustness in dynamic scenes.
We further combine our context encoding with several SLAM backbones. Context+ORB-SLAM3 enables dual-context tracking, but its accuracy remains limited, especially in the extra context (RE 26.93 px, ARE 6.29$^\circ$, ATE 0.446 m). This is because ORB features are less robust than Superpoint under illumination changes and fast geometric variations \cite{10297407}.  Context+AirSLAM and Context+sqrtVINS mainly track the extra-context, since visual-inertial motion estimation tends to favor extra features over intra cues. Context+MASt3R improves intra-context tracking, but still fails to maintain reliable extra-context tracking due to motion blur and unstable external feature extraction.

For AR-HUD baselines, world-anchored overlays \cite{jansen2026mirage} content onto scene objects and thus lose tracking under occlusion in both contexts, whereas screen-fixed HUD (fixed on the AR screen) \cite{8466859} remains visible but can occlude the driver’s view. In addition, both HUD baselines provide positional anchoring without explicit orientation control, limiting their ability to adapt pose between driver and overlays.

In contrast, SEER-VAR consistently tracks both intra- and extra-vehicle contexts with RE of 13.74/17.63 px, ARE of 1.85$^\circ$/5.64$^\circ$, and ATE of 0.054/0.312 m for intra/extra contexts, respectively. This demonstrates that combining context-aware decomposition and CASB is necessary for stable egocentric AR anchoring in driving.

\subsection{Qualitative User Study Results}


\begin{figure}[t]
      \centering
      \includegraphics[width=0.45\textwidth]{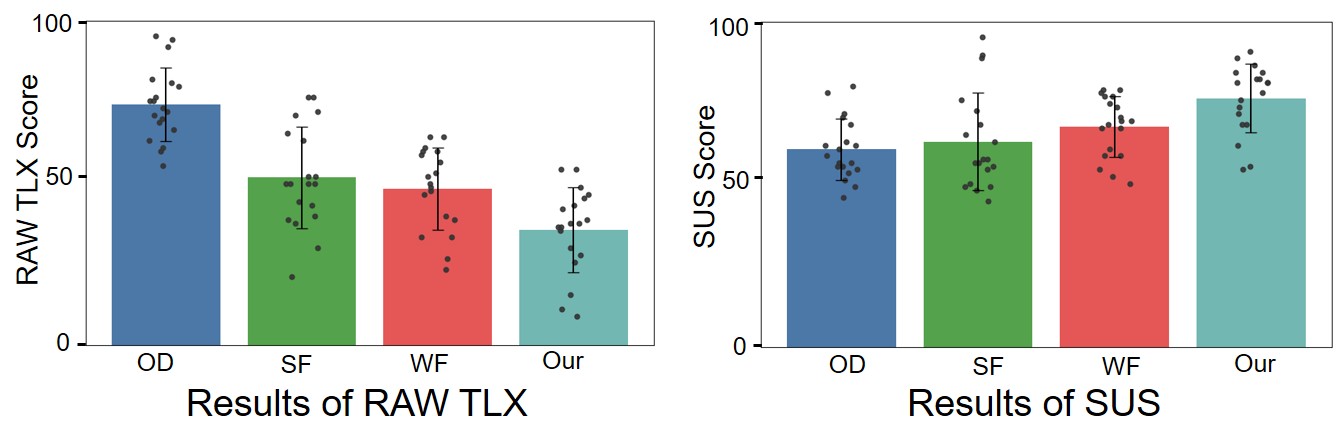}
      \vspace{-10pt}
      \caption{
      Results of Workload Assessment and Usability.} 
      \label{userstudy}
\vspace{-10pt}
\end{figure} %

\begin{figure*}[tb]
  \centering
    \includegraphics[width=0.8\textwidth]{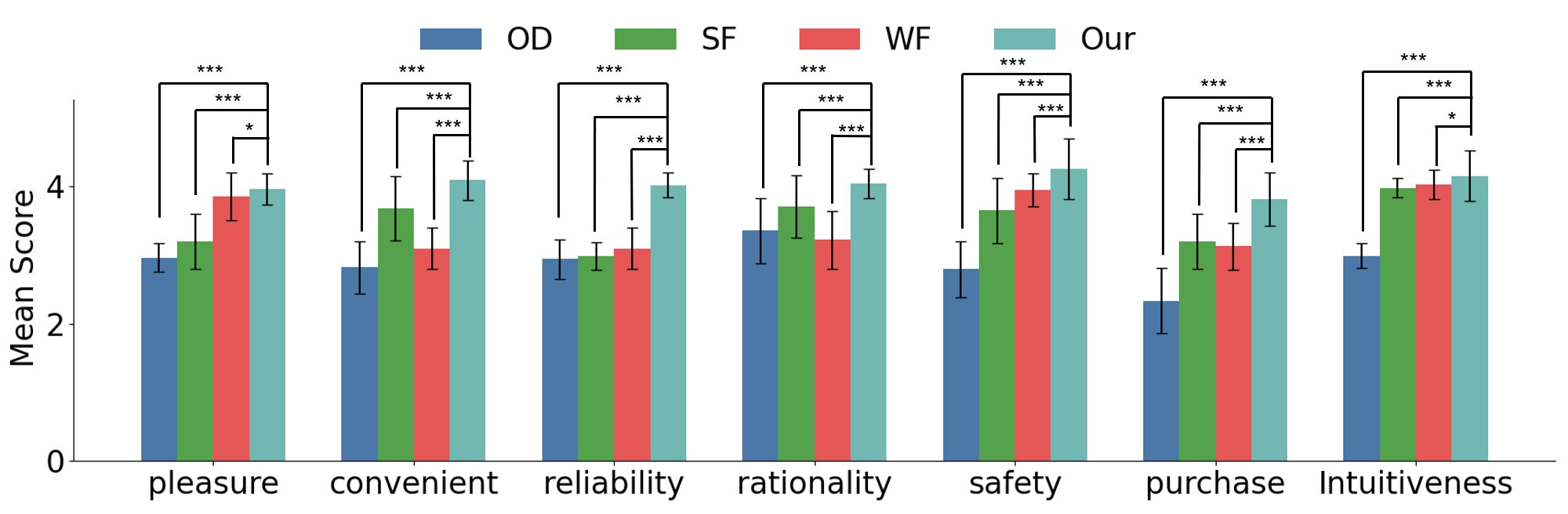}
    \vspace{-5pt}
    \caption{ User study results comparing SEER-VAR vs. AR-HUD Baseline. Asterisks denote significance levels with Bonferroni correction: *p$<$0.05, ***p$<$0.001}
    \label{Questionnaire}
 \vspace{-10pt}
\end{figure*}

We conducted a user study on a driving simulator to compare SEER-VAR with AR-HUD baselines. Overlays were rendered on the HoloLens2 for all methods. We recruited 152 participants from the university (92 male, 60 female). Each participant experienced four methods in randomized order: OD (original dashboard), SF (screen-fixed AR HUD), WF (world-anchored AR HUD), and Our (SEER-VAR). Additional results are shown in Supp. Sec. 19.

After completing all baseline and proposed methods, participants completed a NASA-TLX workload assessment and a System Usability Scale (SUS). The results are shown in \cref{userstudy}. For NASA-TLX results, a Shapiro–Wilk test confirms that all groups follow a normal distribution ($p>0.05$). 
ANOVA analysis finds significant differences between the groups ($F(3,453)=48.86$,$p<0.05$). Post-hoc Tukey HSD analysis shows that all AR methods yielded significantly lower NASA-TLX than OD (minimum difference = $20.36$, 
$p<0.05$), and our method achieves significantly lower scores than the other two AR-HUD baselines(minimum difference = $11.47$, $p<0.05$). Results show that AR-HUD designs reduce workload compared to the original dashboard, and SEER-VAR yields the largest reduction, suggesting that it imposes the least cognitive demand during driving-related tasks. For SUS results, a Shapiro–Wilk test confirms that all
groups follow the normal distribution ($p>0.05$) except SF ($p = 0.03 < 0.05$). ANOVA found significant differences between the groups ($F(3,453)=15.91$,$p < 0.05$). A non-parametric Friedman test confirms the same conclusion. Post-hoc Tukey HSD
analysis shows that our method is significantly higher than
other baselines (minimum difference = $8.11$, $p<0.05$). This suggests that participants perceive SEER-VAR as the most usable and practical interface among the baselines. Qualitatively, participants frequently reported that SEER-VAR remains comfortable to use because the overlay stays front-facing with head motion and appears to float in space without blocking critical road content. In contrast, SF was often described as distracting: as it stays fixed in the view, it continues to occlude the user’s view even when the head turns, potentially covering important roadway information.

In addition, we conducted a survey in which users evaluated all methods in multiple dimensions using a 5-point Likert scale. \cref{Questionnaire} shows the result. Overall, SEER-VAR consistently receives the highest mean scores across pleasure, convenience, reliability, rationality, safety, purchase intention, and intuitiveness. Participants particularly emphasized (i) reduced time to look down, (ii) stable and well-oriented overlays during head motion, and (iii) the benefit of the recommendation module, which automatically selects relevant content rather than displaying persistent UI elements.

\begin{table}[t]
\centering
\caption{Ablation study of SEER-VAR framework under different conditions. “--” denotes that this metric cannot be computed.}
\footnotesize
\setlength{\tabcolsep}{1pt} 
\label{tab:ablation}
\begin{tabular}{lcccccc}

\hline
\multirow{2}{*}{Variants} &
 \multirow{2}{*}{\shortstack{RE\\Intra $\downarrow$} }  &
 \multirow{2}{*}{\shortstack{RE\\Extra $\downarrow$}  } &  
 \multirow{2}{*}{\shortstack{ARE\\Intra $\downarrow$} }  &
 \multirow{2}{*}{\shortstack{ARE\\Extra $\downarrow$}  } &  
 \multirow{2}{*}{\shortstack{ATE\\Intra $\downarrow$} }  &
 \multirow{2}{*}{\shortstack{ATE\\Extra $\downarrow$}  } \\ \\
\hline
 \multirow{1}{*}{\shortstack{w.o. Dynamic mask} }       & 17.72   & 55.16   & 2.81   & 9.32 & 0.102   & 1.847    \\
\multirow{1}{*}{\shortstack{Fixed depth threshold  } }  & 66.53 & -- & 9.76 & -- & 4.53 & --   \\
 \multirow{1}{*}{\shortstack{Bounding box   } }      & 14.06   & 19.82   & 1.94   & 6.05 & 0.059   & 0.336    \\

\hline
\textbf{Ours}     &\textbf{13.74} & \textbf{17.63}  & \textbf{1.85}& \textbf{5.64}& \textbf{0.054}& \textbf{0.312}    \\
\hline
\end{tabular}
\footnotesize{ }
\vspace{-10pt}
\end{table}

\begin{table}[t]
\centering
 \footnotesize
\caption{Ablation of single-context branch selection. Correct branch selections / total images are reported.} 
\label{tab:context_checking_ablation}
\setlength{\tabcolsep}{1pt}
\begin{tabular}{l|cccc|c}
\hline
\multirow{2}{*}{Variants}  & \multicolumn{4}{c|}{Extra-Dominated} & {Intra-Dominated} \\
  & Traffic & Tunnel & Rural & Urban & Dashboard \\
\hline
w.o. Context Checking 
& 0/30 & 8/30 & 13/30 & 18/30 & 10/30 \\
Our 
& 29/30 & 28/30 & 30/30 & 30/30 & 28/30 \\
\hline
\end{tabular}
\vspace{-15pt}
\end{table}

\subsection{Ablation Studies}

We conduct an ablation study to evaluate the impact of each CASB component (\cref{tab:ablation}). Removing dynamic object masks preserves tracking, but increases RE, ARE, and ATE due to moving distractors, especially outdoors where cars and pedestrians dominate feature-rich regions. For each ablation run of a fixed depth threshold, the threshold is set to a percentile
   of the maximum depth value. However, these fixed thresholds lead to complete failure, as relative depth estimates are not consistent. When we substitute segmentation with bounding boxes, SLAM survives, but accuracy suffers, especially inside the vehicle, where boxes over the driver's hands may occlude essential static features. In contrast, our full system maintains robust tracking and achieves the lowest RE, ARE, and ATE, demonstrating the effectiveness of semantic context-aware scene decomposition.

To evaluate the semantic context check module, we sample 30 single-context images from each dominated scenario, including dense traffic, low-light tunnel, rural road, urban road, and dashboard views. As shown in \cref{tab:context_checking_ablation}, the baseline without context checking selects the branch based on the intra- or extra-mask area ratio after depth separation, which becomes unreliable when the depth distribution is not clearly bimodal. In contrast, our semantic context check uses detected representative object categories to determine the dominated context before applying depth splitting, improving branch selection accuracy from 49/150 to 145/150.

\section{Conclusion}
In this work, we present \textbf{SEER-VAR}, an egocentric framework for vehicle-based AR that unifies semantic decomposition, CASB, and GPT-driven contextual recommendations in a single pipeline. SEER-VAR explicitly decouples cabin and road scenes via depth-guided semantic segmentation to enable robust, real-time localization in multi-context environments, and uses LLMs to generate adaptive, spatially grounded AR overlays. We also introduce \textbf{EgoSLAM-Drive}, a real-world benchmark for egocentric SLAM and AR. Experiments show consistent gains over prior methods, and we plan to incorporate richer multimodal cues and stronger reasoning for improved robustness.

{
    \small
    \bibliographystyle{ieeenat_fullname}
    \bibliography{main}

@String(CVPR= {IEEE Conf. Comput. Vis. Pattern Recog.})

@String(AAAI = {AAAI})

@String(CVPRW= {IEEE Conf. Comput. Vis. Pattern Recog. Worksh.})

@String(CVPR  = {CVPR})

@String(CVPRW= {CVPRW})

@inproceedings{detone2018superpoint,
  title={Superpoint: Self-supervised interest point detection and description},
  author={DeTone, Daniel and Malisiewicz, Tomasz and Rabinovich, Andrew},
  booktitle={Proceedings of the IEEE conference on computer vision and pattern recognition workshops},
  pages={224--236},
  year={2018}
}

@inproceedings{jansen2026mirage,
  title={MIRAGE: Enabling Real-Time Automotive Mediated Reality},
  author={Jansen, Pascal and Britten, Julian and Colley, Mark and Sasalovici, Markus and Rukzio, Enrico},
  booktitle={Proceedings of the 2026 CHI Conference on Human Factors in Computing Systems},
  pages={1--20},
  year={2026}
}

@article{narzt2006augmented,
  title={Augmented reality navigation systems},
  author={Narzt, Wolfgang and Pomberger, Gustav and Ferscha, Alois and Kolb, Dieter and M{\"u}ller, Reiner and Wieghardt, Jan and H{\"o}rtner, Horst and Lindinger, Christopher},
  journal={Universal Access in the Information Society},
  volume={4},
  number={3},
  pages={177--187},
  year={2006},
  publisher={Springer}
}

@article{zhou2025edgesam,
  title={EdgeSAM: Prompt-in-the-loop distillation for SAM},
  author={Zhou, Chong and Li, Xiangtai and Loy, Chen Change and Dai, Bo},
  journal={International Journal of Computer Vision},
  volume={133},
  number={12},
  pages={8452--8468},
  year={2025},
  publisher={Springer}
}

@article{yolo26e_ultralytics,
  title={Ultralytics {YOLO26}: Unified real-time end-to-end vision models},
  author={Jocher, Glenn and Qiu, Jing and Liu, Mengyu and Lyu, Shuai and Akyon, Fatih Cagatay and Kalfaoglu, Muhammet Esat},
  journal={arXiv preprint arXiv:2606.03748},
  year={2026}
}

@article{pomarjanschi2012gaze,
  title={Gaze guidance reduces the number of collisions with pedestrians in a driving simulator},
  author={Pomarjanschi, Laura and Dorr, Michael and Barth, Erhardt},
  journal={ACM Transactions on Interactive Intelligent Systems (TiiS)},
  volume={1},
  number={2},
  pages={1--14},
  year={2012},
  publisher={ACM New York, NY, USA}
}

@INPROCEEDINGS{4538831,
  author={Tonnis, Marcus and Lange, Christian and Klinker, Gudrun},
  booktitle={2007 6th IEEE and ACM International Symposium on Mixed and Augmented Reality}, 
  title={Visual Longitudinal and Lateral Driving Assistance in the Head-Up Display of Cars}, 
  year={2007},
  volume={},
  number={},
  pages={91-94},
  doi={10.1109/ISMAR.2007.4538831}}

@INPROCEEDINGS{7789504,
  author={Alletto, Stefano and Palazzi, Andrea and Solera, Francesco and Calderara, Simone and Cucchiara, Rita},
  booktitle={2016 IEEE Conference on Computer Vision and Pattern Recognition Workshops (CVPRW)}, 
  title={DR(eye)VE: A Dataset for Attention-Based Tasks with Applications to Autonomous and Assisted Driving}, 
  year={2016},
  volume={},
  number={},
  pages={54-60},
  doi={10.1109/CVPRW.2016.14}}

@inproceedings{akbiyik2023routeformer,
     title={Leveraging Driver Field-of-View for Multimodal Ego-Trajectory Prediction},
  author={Akbiyik, M Eren and Savov, Nedko and Paudel, Danda Pani and Popovic, Nikola and Vater, Christian and Hilliges, Otmar and Van Gool, Luc and Wang, Xi},
    booktitle={International Conference on Learning Representations},
    year={2025}
}

@INPROCEEDINGS{8613746,
  author={Runz, Martin and Buffier, Maud and Agapito, Lourdes},
  booktitle={2018 IEEE International Symposium on Mixed and Augmented Reality (ISMAR)}, 
  title={MaskFusion: Real-Time Recognition, Tracking and Reconstruction of Multiple Moving Objects}, 
  year={2018},
  volume={},
  number={},
  pages={10-20},
  doi={10.1109/ISMAR.2018.00024}}

@ARTICLE{8466859,
  author={Merenda, Coleman and Kim, Hyungil and Tanous, Kyle and Gabbard, Joseph L. and Feichtl, Blake and Misu, Teruhisa and Suga, Chihiro},
  journal={IEEE Transactions on Visualization and Computer Graphics}, 
  title={Augmented Reality Interface Design Approaches for Goal-directed and Stimulus-driven Driving Tasks}, 
  year={2018},
  volume={24},
  number={11},
  pages={2875-2885},
  doi={10.1109/TVCG.2018.2868531}}

@Article{Peng2025TRO,
  title = {{$\sqrt{\mathbf{VINS}}$}: Robust and Ultrafast Square-Root Filter-Based {3D} Motion Tracking},
  author  = {Peng, Yuxiang and Chen, Chuchu and Wu, Kejian and Huang, Guoquan},
  journal = {IEEE Transactions on Robotics},
  volume  = {41},
  pages   = {6570--6589},
  year    = {2025},
  doi     = {10.1109/TRO.2025.3626607}
}

@article{Geiger2013IJRR,
  title   = {Vision Meets Robotics: The {KITTI} Dataset},
  author  = {Geiger, Andreas and Lenz, Philip and Stiller, Christoph and Urtasun, Raquel},
  journal = {The International Journal of Robotics Research},
  volume  = {32},
  number  = {11},
  pages   = {1231--1237},
  year    = {2013},
  doi     = {10.1177/0278364913491297}
}

@inproceedings{li2024cto,
  title={CTO-SLAM: contour tracking for object-level robust 4D SLAM},
  author={Li, Xiaohan and Liu, Dong and Wu, Jun},
  booktitle={Proceedings of the AAAI Conference on Artificial Intelligence},
  volume={38},
  pages={10323--10331},
  year={2024}
}

@inproceedings{fan2024schurvins,
  title={Schurvins: Schur complement-based lightweight visual inertial navigation system},
  author={Fan, Yunfei and Zhao, Tianyu and Wang, Guidong},
  booktitle={Proceedings of the IEEE/CVF Conference on Computer Vision and Pattern Recognition},
  pages={17964--17973},
  year={2024}
}

@article{xu2025airslam,
  title   = {{AirSLAM}: An Efficient and Illumination-Robust Point-Line Visual {SLAM} System},
  author  = {Xu, Kuan and Hao, Yuefan and Yuan, Shenghai and Wang, Chen and Xie, Lihua},
  journal = {IEEE Transactions on Robotics},
  volume  = {41},
  pages   = {1673--1692},
  year    = {2025},
  doi     = {10.1109/TRO.2025.3539171}
}

@inproceedings{brohan2023can,
 title     = {Do As {I} Can, Not As {I} Say: Grounding Language in Robotic Affordances},
  author    = {Ichter, Brian and
               Brohan, Anthony and
               Chebotar, Yevgen and
               Finn, Chelsea and
               Hausman, Karol and
               Herzog, Alexander and
               Ho, Daniel and
               Ibarz, Julian and
               Irpan, Alex and
               Jang, Eric and
               Julian, Ryan and
               Kalashnikov, Dmitry and
               Levine, Sergey and
               Lu, Yao and
               Parada, Carolina and
               Rao, Kanishka and
               Sermanet, Pierre and
               Toshev, Alexander T and
               Vanhoucke, Vincent and
               Xia, Fei and
               Xiao, Ted and
               Xu, Peng and
               Yan, Mengyuan and
               Brown, Noah and
               Ahn, Michael and
               Cortes, Omar and
               Sievers, Nicolas and
               Tan, Clayton and
               Xu, Sichun and
               Reyes, Diego and
               Rettinghouse, Jarek and
               Quiambao, Jornell and
               Pastor, Peter and
               Luu, Linda and
               Lee, Kuang-Huei and
               Kuang, Yuheng and
               Jesmonth, Sally and
               Joshi, Nikhil J. and
               Jeffrey, Kyle and
               Ruano, Rosario Jauregui and
               Hsu, Jasmine and
               Gopalakrishnan, Keerthana and
               David, Byron and
               Zeng, Andy and
               Fu, Chuyuan Kelly},
  booktitle = {Proceedings of the 6th Conference on Robot Learning},
  series    = {Proceedings of Machine Learning Research},
  volume    = {205},
  pages     = {287--318},
  publisher = {PMLR},
  year      = {2023}
}

@inproceedings{liu2022egocentric,
  title={Egocentric Activity Recognition and Localization on a 3D Map},
  author={Liu, Miao and Ma, Lingni and Somasundaram, Kiran and Li, Yin and Grauman, Kristen and Rehg, James M and Li, Chao},
  booktitle={European Conference on Computer Vision},
  pages={621--638},
  year={2022}
}

@INPROCEEDINGS{8658783,
  author={Patra, Suvam and Gupta, Kartikeya and Ahmad, Faran and Arora, Chetan and Banerjee, Subhashis},
  booktitle={2019 IEEE Winter Conference on Applications of Computer Vision (WACV)}, 
  title={EGO-SLAM: A Robust Monocular SLAM for Egocentric Videos}, 
  year={2019},
  volume={},
  number={},
  pages={31-40},
  doi={10.1109/WACV.2019.00011}}

@inproceedings{murai2025mast3r,
  title={MASt3R-SLAM: Real-time dense SLAM with 3D reconstruction priors},
  author={Murai, Riku and Dexheimer, Eric and Davison, Andrew J},
  booktitle={Proceedings of the Computer Vision and Pattern Recognition Conference},
  pages={16695--16705},
  year={2025}
}

@INPROCEEDINGS{8575524,
  title     = {{Mask-SLAM}: Robust Feature-Based Monocular {SLAM} by Masking Using Semantic Segmentation},
  author    = {Kaneko, Masaya and Iwami, Kazuya and Ogawa, Toru and Yamasaki, Toshihiko and Aizawa, Kiyoharu},
  booktitle = {Proceedings of the IEEE Conference on Computer Vision and Pattern Recognition (CVPR) Workshops},
  pages     = {258--266},
  month     = jun,
  year      = {2018}
}

@article{haselberger2024situation,
  title   = {Situation Awareness for Driver-Centric Driving Style Adaptation},
  author  = {Haselberger, Johann and Stuhr, Bonifaz and Schick, Bernhard and M{\"u}ller, Steffen},
  journal = {IEEE Transactions on Intelligent Vehicles},
  volume  = {10},
  number  = {3},
  pages   = {1959--1973},
  year    = {2025},
  doi     = {10.1109/TIV.2024.3440634}
}

@article{riener2018special,
  title   = {Special Issue of {Presence}: Virtual and Augmented Reality Virtual and Augmented Reality for Autonomous Driving and Intelligent Vehicles: Guest Editors' Introduction},
  author  = {Riener, Andreas and Gabbard, Joe and Trivedi, Mohan},
  journal = {Presence: Teleoperators and Virtual Environments},
  volume  = {27},
  number  = {1},
  pages   = {i--iv},
  year    = {2018},
  doi     = {10.1162/pres_e_00323}
}

@article{gabbard2014behind,
  title={Behind the glass: Driver challenges and opportunities for AR automotive applications},
  author={Gabbard, Joseph L and Fitch, Gregory M and Kim, Hyungil},
  journal={Proceedings of the IEEE},
  volume={102},
  number={2},
  pages={124--136},
  year={2014},
  publisher={IEEE}
}

@inproceedings{grauman2022ego4d,
  title={Ego4d: Around the world in 3,000 hours of egocentric video},
  author={Grauman, Kristen and Westbury, Andrew and Byrne, Eugene and Chavis, Zachary and Furnari, Antonino and Girdhar, Rohit and Hamburger, Jackson and Jiang, Hao and Liu, Miao and Liu, Xingyu and others},
  booktitle={Proceedings of the IEEE/CVF conference on computer vision and pattern recognition},
  pages={18995--19012},
  year={2022}
}

@inproceedings{shridhar2022cliport,
  title={Cliport: What and where pathways for robotic manipulation},
  author={Shridhar, Mohit and Manuelli, Lucas and Fox, Dieter},
  booktitle={Conference on robot learning},
  pages={894--906},
  year={2022},
  organization={PMLR}
}

@article{xu2022seer,
  title={Self-supervised pretraining of visual features in the wild},
  author={Goyal, Priya and Caron, Mathilde and Lefaudeux, Benjamin and Xu, Min and Wang, Pengchao and Pai, Vivek and Singh, Mannat and Liptchinsky, Vitaliy and Misra, Ishan and Joulin, Armand and Bojanowski, Piotr},
  journal={arXiv preprint arXiv:2103.01988},
  year={2021}
}

@inproceedings{alayrac2022flamingo,
  author       = {Alayrac, Jean‑Baptiste and Donahue, Jeff and Luc, Pauline and Miech, Antoine and Barr, Iain and Hasson, Yana and Lenc, Karel and Mensch, Arthur and Millican, Katie and Reynolds, Malcolm and Ring, Roman and Rutherford, Eliza and Cabi, Serkan and Han, Tengda and Gong, Zhitao and Samangooei, Sina and Monteiro, Marianne and Menick, Jacob L. and Borgeaud, Sebastian and Brock, Andrew and Nematzadeh, Aida and Sharifzadeh, Sahand and Binkowski, Mikolaj and Barreira, Ricardo and Vinyals, Oriol and Zisserman, Andrew and Simonyan, Karen},
  title        = {Flamingo: a Visual Language Model for Few‑Shot Learning},
  booktitle    = {Advances in Neural Information Processing Systems 35 (NeurIPS 2022)},
  year         = {2022},
  publisher    = {NeurIPS / Curran Associates},
  address      = {New Orleans, LA, USA},
}

@article{teed2021droid,
  title={Droid-slam: Deep visual slam for monocular, stereo, and rgb-d cameras},
  author={Teed, Zachary and Deng, Jia},
  journal={Advances in neural information processing systems},
  volume={34},
  pages={16558--16569},
  year={2021}
}

@article{damen2022rescaling,
  title   = {Rescaling Egocentric Vision: Collection, Pipeline and Challenges for {EPIC-KITCHENS-100}},
  author  = {Damen, Dima and Doughty, Hazel and Farinella, Giovanni Maria and Furnari, Antonino and Kazakos, Evangelos and Ma, Jian and Moltisanti, Davide and Munro, Jonathan and Perrett, Toby and Price, Will and Wray, Michael},
  journal = {International Journal of Computer Vision},
  volume  = {130},
  number  = {1},
  pages   = {33--55},
  year    = {2022},
  doi     = {10.1007/s11263-021-01531-2}
}

@ARTICLE{ORBSLAM3_TRO,
  author={Campos, Carlos and Elvira, Richard and Rodríguez, Juan J. Gómez and M. Montiel, José M. and D. Tardós, Juan},
  journal={IEEE Transactions on Robotics}, 
  title={ORB-SLAM3: An Accurate Open-Source Library for Visual, Visual–Inertial, and Multimap SLAM}, 
  year={2021},
  volume={37},
  number={6},
  pages={1874-1890},
  doi={10.1109/TRO.2021.3075644}}

@article{fischler1981random,
  title={Random sample consensus: a paradigm for model fitting with applications to image analysis and automated cartography},
  author={Fischler, Martin A and Bolles, Robert C},
  journal={Communications of the ACM},
  volume={24},
  number={6},
  pages={381--395},
  year={1981},
  publisher={ACM New York, NY, USA}
}

@article{yang2024depth,
  title={Depth anything v2},
  author={Yang, Lihe and Kang, Bingyi and Huang, Zilong and Zhao, Zhen and Xu, Xiaogang and Feng, Jiashi and Zhao, Hengshuang},
  journal={Advances in Neural Information Processing Systems},
  volume={37},
  pages={21875--21911},
  year={2024}
}

@article{mur2015orb,
  title={ORB-SLAM: A versatile and accurate monocular SLAM system},
  author={Mur-Artal, Raul and Montiel, Jose Maria Martinez and Tardos, Juan D},
  journal={IEEE transactions on robotics},
  volume={31},
  number={5},
  pages={1147--1163},
  year={2015},
  publisher={IEEE}
}

@inproceedings{GroundingDINO,
  title={Grounding DINO: Marrying DINO with Grounded Pre-training for Open-Set Object Detection},
  author={Liu, Shilong and Zeng, Zhaoyang and Ren, Tianhe and Li, Feng and Zhang, Hao and Yang, Jie and Jiang, Qing and Li, Chunyuan and Yang, Jianwei and Su, Hang and others},
  booktitle={European Conference on Computer Vision},
  pages={38--55},
  year={2024}
}

@inproceedings{caesar2020nuscenes,
  title={nuscenes: A multimodal dataset for autonomous driving},
  author={Caesar, Holger and Bankiti, Varun and Lang, Alex H and Vora, Sourabh and Liong, Venice Erin and Xu, Qiang and Krishnan, Anush and Pan, Yu and Baldan, Giancarlo and Beijbom, Oscar},
  booktitle={Proceedings of the IEEE/CVF conference on computer vision and pattern recognition},
  pages={11621--11631},
  year={2020}
}

@inproceedings{sun2020scalability,
  title={Scalability in perception for autonomous driving: Waymo open dataset},
  author={Sun, Pei and Kretzschmar, Henrik and Dotiwalla, Xerxes and Chouard, Aurelien and Patnaik, Vijaysai and Tsui, Paul and Guo, James and Zhou, Yin and Chai, Yuning and Caine, Benjamin and others},
  booktitle={Proceedings of the IEEE/CVF conference on computer vision and pattern recognition},
  pages={2446--2454},
  year={2020}
}

@inproceedings{brizi2024vbr,
  title={Vbr: A vision benchmark in rome},
  author={Brizi, Leonardo and Giacomini, Emanuele and Di Giammarino, Luca and Ferrari, Simone and Salem, Omar and De Rebotti, Lorenzo and Grisetti, Giorgio},
  booktitle={2024 IEEE International Conference on Robotics and Automation (ICRA)},
  pages={15868--15874},
  year={2024},
  organization={IEEE}
}

@article{lv2024aria,
  title={Aria everyday activities dataset},
  author={Lv, Zhaoyang and Charron, Nicholas and Moulon, Pierre and Gamino, Alexander and Peng, Cheng and Sweeney, Chris and Miller, Edward and Tang, Huixuan and Meissner, Jeff and Dong, Jing and others},
  journal={arXiv preprint arXiv:2402.13349},
  year={2024}
}

@article{zhang2022hilti,
  title   = {{Hilti-Oxford} Dataset: A Millimeter-Accurate Benchmark for Simultaneous Localization and Mapping},
  author  = {Zhang, Lintong and Helmberger, Michael and Fu, Lanke Frank Tarimo and Wisth, David and Camurri, Marco and Scaramuzza, Davide and Fallon, Maurice},
  journal = {IEEE Robotics and Automation Letters},
  volume  = {8},
  number  = {1},
  pages   = {408--415},
  year    = {2023},
  doi     = {10.1109/LRA.2022.3226077}
}

@article{kong2025aria,
  title={Aria Gen 2 Pilot Dataset},
  author={Kong, Chen and Fort, James and Kang, Aria and Wittmer, Jonathan and Green, Simon and Shen, Tianwei and Zhao, Yipu and Peng, Cheng and Solaira, Gustavo and Berkovich, Andrew and others},
  journal={arXiv preprint arXiv:2510.16134},
  year={2025}
}

@InProceedings{Li_2026_CVPR,
    author    = {Li, Peizheng and Zhang, Zhenghao and Holtz, David and Yu, Hang and Yang, Yutong and Lai, Yuzhi and Song, Rui and Geiger, Andreas and Zell, Andreas},
    title     = {SpaceDrive: Infusing Spatial Awareness into VLM-based Autonomous Driving},
    booktitle = {Proceedings of the IEEE/CVF Conference on Computer Vision and Pattern Recognition (CVPR)},
    month     = {June},
    year      = {2026},
    pages     = {40096-40107}
}

@article{lai2026objectselectionmarkerlessgazebasedrobot,
  title   = {Beyond Object Selection: Markerless Gaze-based Robot Placement at Arbitrary Positions},
  author  = {Lai, Yuzhi and Marx, William and Yuan, Shenghai and Li, Peizheng and Ran, Zhuoyu and Zell, Andreas},
  journal = {arXiv preprint arXiv:2609.00478},
  year    = {2026}
}

@inproceedings{sturm2012benchmark,
  title={A benchmark for the evaluation of RGB-D SLAM systems},
  author={Sturm, J{\"u}rgen and Engelhard, Nikolas and Endres, Felix and Burgard, Wolfram and Cremers, Daniel},
  booktitle={2012 IEEE/RSJ international conference on intelligent robots and systems},
  pages={573--580},
  year={2012},
  organization={IEEE}
}

@ARTICLE{11534298,
  author={Lai, Yuzhi and Yuan, Shenghai and Li, Peizheng and Zhang, Boya and Kiefer, Benjamin and Deng, Tianchen and Zell, Andreas},
  journal={IEEE Transactions on Automation Science and Engineering}, 
  title={FAM-HRI: Foundation-Model Assisted Multimodal Human–Robot Interaction Combining Gaze and Speech}, 
  year={2026},
  volume={23},
  number={},
  pages={10521-10534},
  doi={10.1109/TASE.2026.3695438}}

@inproceedings{wang2026seeing,
  title     = {Seeing in the Dark: Benchmarking Egocentric {3D} Vision with the {Oxford Day-and-Night} Dataset},
  author    = {Wang, Zirui and Bian, Wenjing and Li, Xinghui and Tao, Yifu and Wang, Jianeng and Fallon, Maurice and Prisacariu, Victor},
  booktitle = {Advances in Neural Information Processing Systems},
  volume    = {38},
  pages     = {18784--18808},
  publisher = {Curran Associates, Inc.},
  year      = {2025},
  doi       = {10.52202/085713-0561}
}

@INPROCEEDINGS{7989443,
  author={Pfrommer, Bernd and Sanket, Nitin and Daniilidis, Kostas and Cleveland, Jonas},
  booktitle={2017 IEEE International Conference on Robotics and Automation (ICRA)}, 
  title={PennCOSYVIO: A challenging Visual Inertial Odometry benchmark}, 
  year={2017},
  volume={},
  number={},
  pages={3847-3854},
  doi={10.1109/ICRA.2017.7989443}}

@inproceedings{wenzel20204seasons,
  title={4Seasons: A cross-season dataset for multi-weather SLAM in autonomous driving},
  author={Wenzel, Patrick and Wang, Rui and Yang, Nan and Cheng, Qing and Khan, Qadeer and Von Stumberg, Lukas and Zeller, Niclas and Cremers, Daniel},
  booktitle={DAGM German Conference on Pattern Recognition},
  pages={404--417},
  year={2020},
  organization={Springer}
}

@INPROCEEDINGS{10297407,
  title     = {{SP-VSLAM}: Monocular Visual-{SLAM} Algorithm Based on {SuperPoint} Network},
  author    = {Yin, Zhenyu and Feng, Dan and Fan, Chao and Ju, Chengen and Zhang, Feiqing},
  booktitle = {2023 15th International Conference on Communication Software and Networks (ICCSN)},
  pages     = {456--459},
  publisher = {IEEE},
  year      = {2023},
  doi       = {10.1109/ICCSN57992.2023.10297407}
}

@article{openai2023gpt4,
  title={GPT-4 Technical Report},
  author={OpenAI},
  journal={arXiv preprint arXiv:2303.08774},
  year={2023}
}

@misc{openai_realtime_api,
  author       = {{OpenAI}},
  title        = {{GPT-Realtime-2.1}},
  howpublished = {OpenAI API Documentation},
  year         = {2026},
  url          = {https://developers.openai.com/api/docs/models/gpt-realtime-2.1}
}

@inproceedings{krishnan2025benchmarking,
  title={Benchmarking Egocentric Visual-Inertial SLAM at City Scale},
  author={Krishnan, Anusha and Liu, Shaohui and Sarlin, Paul-Edouard and Gentilhomme, Oscar and Caruso, David and Monge, Maurizio and Newcombe, Richard and Engel, Jakob and Pollefeys, Marc},
  booktitle={Proceedings of the IEEE/CVF International Conference on Computer Vision},
  pages={25207--25217},
  year={2025}
}

@inproceedings{kirillov2023sam,
  title={Segment anything},
  author={Kirillov, Alexander and Mintun, Eric and Ravi, Nikhila and Mao, Hanzi and Rolland, Chloe and Gustafson, Laura and Xiao, Tete and Whitehead, Spencer and Berg, Alexander C and Lo, Wan-Yen and others},
  booktitle={Proceedings of the IEEE/CVF international conference on computer vision},
  pages={4015--4026},
  year={2023}
}

@INPROCEEDINGS{9341801,
  author={Wang, Wenshan and Zhu, Delong and Wang, Xiangwei and Hu, Yaoyu and Qiu, Yuheng and Wang, Chen and Hu, Yafei and Kapoor, Ashish and Scherer, Sebastian},
  booktitle={2020 IEEE/RSJ International Conference on Intelligent Robots and Systems (IROS)}, 
  title={TartanAir: A Dataset to Push the Limits of Visual SLAM}, 
  year={2020},
  volume={},
  number={},
  pages={4909-4916},
  doi={10.1109/IROS45743.2020.9341801}}

@article{GroundedSAM,
  title={Grounded sam: Assembling open-world models for diverse visual tasks},
  author={Ren, Tianhe and Liu, Shilong and Zeng, Ailing and Lin, Jing and Li, Kunchang and Cao, He and Chen, Jiayu and Huang, Xinyu and Chen, Yukang and Yan, Feng and others},
  journal={arXiv preprint arXiv:2401.14159},
  year={2024}
}

@ARTICLE{8421015,
  author={Bescos, Berta and Fácil, José M. and Civera, Javier and Neira, José},
  journal={IEEE Robotics and Automation Letters}, 
  title={DynaSLAM: Tracking, Mapping, and Inpainting in Dynamic Scenes}, 
  year={2018},
  volume={3},
  number={4},
  pages={4076-4083},
  doi={10.1109/LRA.2018.2860039}}

@inproceedings{li2024situationadapt,
  title={Situationadapt: Contextual ui optimization in mixed reality with situation awareness via llm reasoning},
  author={Li, Zhipeng and Gebhardt, Christoph and Inglin, Yves and Steck, Nicolas and Streli, Paul and Holz, Christian},
  booktitle={Proceedings of the 37th Annual ACM Symposium on User Interface Software and Technology},
  pages={1--13},
  year={2024}
}

@inproceedings{zaman2023vicarious,
  title={Vicarious: Context-aware viewpoints selection for mixed reality collaboration},
  author={Zaman, Faisal and Anslow, Craig and Rhee, Taehyun James},
  booktitle={Proceedings of the 29th ACM Symposium on Virtual Reality Software and Technology},
  pages={1--11},
  year={2023}
}

@article{2025exploringcontextawarellmdrivenlocomotion,
  title={Exploring Context-aware and {LLM}-driven Locomotion for Immersive Virtual Reality},
  author={{\"O}zdel, S{\"u}leyman and Buldu, Kadir Burak and Kasneci, Enkelejda and Bozkir, Efe},
  journal={Proceedings of the ACM on Human-Computer Interaction},
  volume={10},
  number={5},
  articleno={MHCI6433},
  pages={MHCI6433:1--MHCI6433:28},
  numpages={28},
  year={2026},
  month=aug,
  publisher={Association for Computing Machinery},
  doi={10.1145/3821669}
}

@inproceedings{huang2022innermonologue,
  title     = {Inner Monologue: Embodied Reasoning through Planning with Language Models},
  author    = {Huang, Wenlong and
               Xia, Fei and
               Xiao, Ted and
               Chan, Harris and
               Liang, Jacky and
               Florence, Pete and
               Zeng, Andy and
               Tompson, Jonathan and
               Mordatch, Igor and
               Chebotar, Yevgen and
               Sermanet, Pierre and
               Jackson, Tomas and
               Brown, Noah and
               Luu, Linda and
               Levine, Sergey and
               Hausman, Karol and
               Ichter, Brian},
  booktitle = {Proceedings of the 6th Conference on Robot Learning},
  series    = {Proceedings of Machine Learning Research},
  volume    = {205},
  pages     = {1769--1782},
  publisher = {PMLR},
  year      = {2023}
}

@inproceedings{xu2022arshopping,
  title={ARShopping: In-store shopping decision support through augmented reality and immersive visualization},
  author={Xu, Bingjie and Guo, Shunan and Koh, Eunyee and Hoffswell, Jane and Rossi, Ryan and Du, Fan},
  booktitle={2022 IEEE Visualization and Visual Analytics (VIS)},
  pages={120--124},
  year={2022},
  organization={IEEE}
}

@inproceedings{an2021arshoe,
  title={ARShoe: Real-time augmented reality shoe try-on system on smartphones},
  author={An, Shan and Che, Guangfu and Guo, Jinghao and Zhu, Haogang and Ye, Junjie and Zhou, Fangru and Zhu, Zhaoqi and Wei, Dong and Liu, Aishan and Zhang, Wei},
  booktitle={Proceedings of the 29th ACM International Conference on Multimedia},
  pages={1111--1119},
  year={2021}
}

@article{xu2025enhancing,
  title   = {Enhancing Scene Coordinate Regression with Efficient Keypoint Detection and Sequential Information},
  author  = {Xu, Kuan and Jiang, Zeyu and Cao, Haozhi and Yuan, Shenghai and Wang, Chen and Xie, Lihua},
  journal = {IEEE Robotics and Automation Letters},
  volume  = {10},
  number  = {10},
  pages   = {9932--9939},
  year    = {2025},
  doi     = {10.1109/LRA.2025.3598670}
}

@article{shi2025fashionpose,
  title={{FashionPose}: Text to pose to relight image generation for personalized fashion visualization},
  author={Shi, Chuancheng and Chen, Yixiang and Lei, Burong and Chen, Jichao},
  journal={arXiv preprint arXiv:2507.13311},
  year={2025}
}

@inproceedings{shi2026sgmhand,
  title={{SGMHand}: Structure-Guided Modulation for Structure-Aware Hand Inpainting},
  author={Shi, Chuancheng and Guo, Shiming and Shui, Ke and Chen, Yixiang and Shen, Fei},
  booktitle={Proceedings of the AAAI Conference on Artificial Intelligence},
  volume={40},
  number={11},
  pages={8942--8950},
  year={2026}
}

@article{shi2026correct,
  title={Do correct answers have stable latent knowledge support?},
  author={Shi, Chuancheng and Li, Shangze and Wu, Wenhua and Hu, Kun and Shen, Fei and Wang, Zhiyong and Chua, Tat-Seng},
  journal={arXiv preprint arXiv:2610.15277},
  year={2026}
}

@article{wang2026cruise,
  title={{CRUISE}: Vision-Language Model-Guided Uncertainty-Aware Cross-Modal Sensor Fusion for Robust Autonomous Driving},
  author={Wang, Junyao and Xu, Yulin and Li, Yu and Khargonekar, Pramod and Faruque, Mohammad Abdullah Al},
  journal={arXiv preprint arXiv:2608.09202},
  year={2026}
}
}

\end{document}